\documentclass[conference]{IEEEtran}
\IEEEoverridecommandlockouts
\usepackage{cite}
\usepackage{amsmath,amssymb,amsfonts}
\usepackage{algorithmic}
\usepackage{graphicx}
\usepackage{textcomp}
\usepackage{array}
\newcolumntype{M}[1]{>{\centering\arraybackslash}p{#1}}
\usepackage{enumitem}
\usepackage{hyperref}
\usepackage{xcolor}
\usepackage{color, soul}
\usepackage{etoolbox}
\makeatletter
\patchcmd{\@makecaption}
  {\scshape}
  {}
  {}
  {}
\makeatother
\def\BibTeX{{\rm B\kern-.05em{\sc i\kern-.025em b}\kern-.08em
    T\kern-.1667em\lower.7ex\hbox{E}\kern-.125emX}}
\begin{document}

\title{STRIDE : Scene Text Recognition In-Device}

\makeatletter
\newcommand{\linebreakand}{%
  \end{@IEEEauthorhalign}
  \hfill\mbox{}\par
  \mbox{}\hfill\begin{@IEEEauthorhalign}
}
\makeatother

\author{\IEEEauthorblockN{Rachit S Munjal}
\IEEEauthorblockA{\textit{On-Device AI} \\
\textit{Samsung R\&D Institute}\\
Bangalore, India\\
rachit.m@samsung.com}
\and
\IEEEauthorblockN{Arun D Prabhu}
\IEEEauthorblockA{\textit{On-Device AI} \\
\textit{Samsung R\&D Institute}\\
Bangalore, India\\
arun.prabhu@samsung.com}
\and
\IEEEauthorblockN{Nikhil Arora}
\IEEEauthorblockA{\textit{On-Device AI} \\
\textit{Samsung R\&D Institute}\\
Bangalore, India\\
n.arora@samsung.com}
 \linebreakand
\IEEEauthorblockN{Sukumar Moharana}
\IEEEauthorblockA{\textit{On-Device AI} \\
\textit{Samsung R\&D Institute}\\
Bangalore, India\\
msukumar@samsung.com}
\and
\IEEEauthorblockN{Gopi Ramena}
\IEEEauthorblockA{\textit{On-Device AI} \\
\textit{Samsung R\&D Institute}\\
Bangalore, India\\
gopi.ramena@samsung.com}}

\maketitle

\begin{abstract}
 Optical Character Recognition (OCR) systems have been widely used in various applications for extracting semantic information from images. To give the user more control over their privacy, an on-device solution is needed. The current state of the art models are too heavy and complex to be deployed on-device. We develop an efficient lightweight scene text recognition (STR) system, which has only 0.88M parameters and performs real-time text recognition. Attention modules tend to boost the accuracy of STR networks but are generally slow and not optimized for device inference. So, we propose the use of convolution attention modules to the text recognition networks, which aims to provide channel and spatial attention information to the LSTM module by adding very minimal computational cost. It boosts our word accuracy on ICDAR 13 dataset by almost 2\%. We also introduce a novel orientation classifier module, to support the simultaneous recognition of both horizontal and vertical text. 
 
 The proposed model surpasses on-device metrics of inference time and memory footprint and achieves comparable accuracy when compared to the leading commercial and other open-source OCR engines. We deploy the system on-device with an inference speed of 2.44 ms per word on the Exynos 990 chipset device and achieve an accuracy of 88.4\% on ICDAR-13 dataset. 
 
\end{abstract}

\section{Introduction}

Mobile phones are ubiquitous today. Billions of images are captured every day for personal and social use. This has driven the need for image understanding tasks such as object detection and optical character recognition (OCR). OCR is one of the most renowned and foremost discussed Computer Vision task, which is used to convert the text in images to electronic form in order to analyze digitized data. The recognition of text has a wide range of applications such as image retrieval, scene understanding, keyword-based image hashing, document analysis, and machine translation. Though OCR pipelines work well to retrieve text from scanned documents, these traditional methods fail to work on images occurring in natural scenes. We focus on analyzing and extracting text from real-world scene images, commonly known as Scene Text Recognition (STR).

\begin{figure}[!b]
\centerline{\includegraphics[scale=0.35]{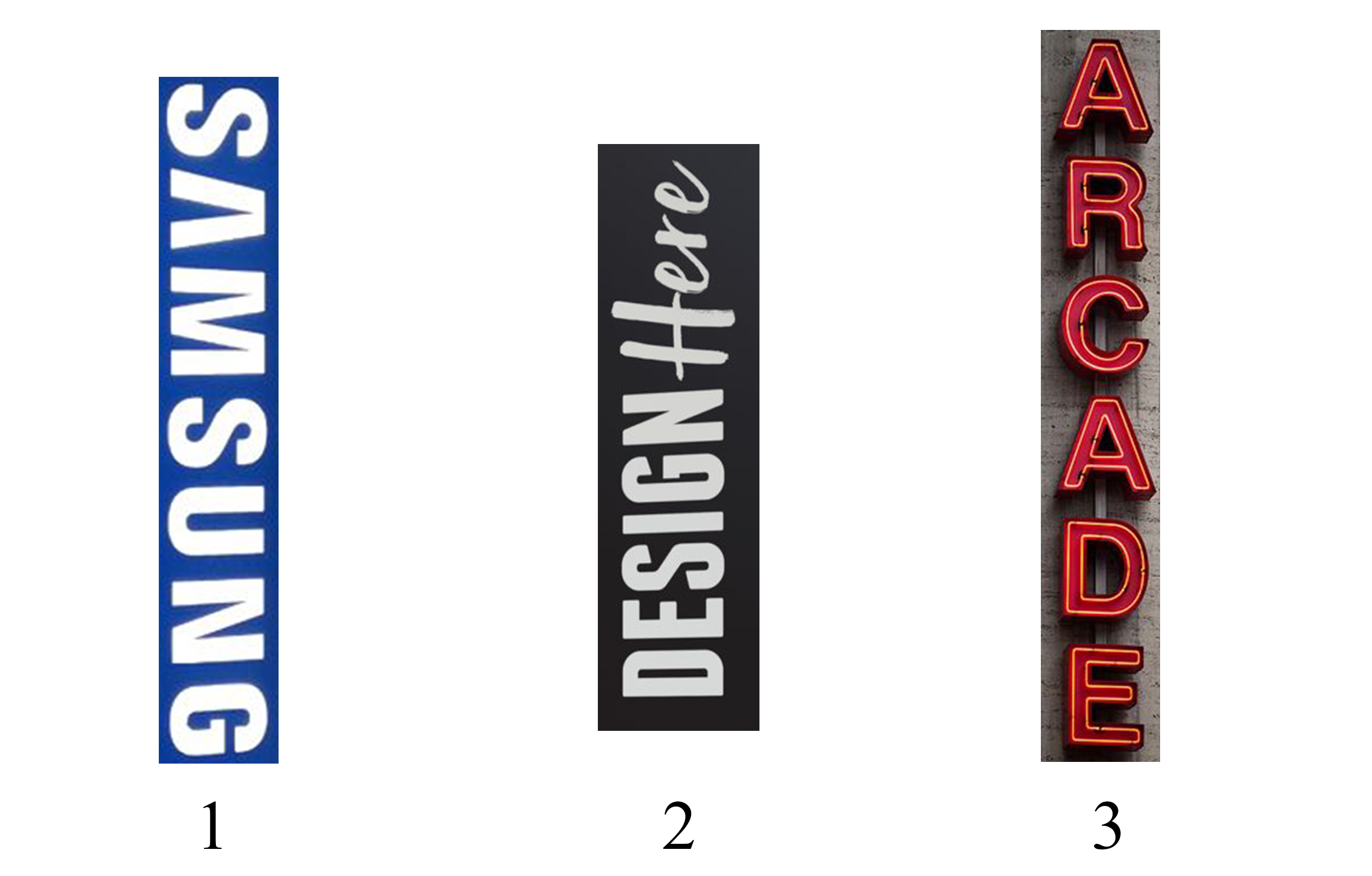}}
\caption{1 and 2 are orthogonally rotated images, whereas 3 is a type of vertical image which is not handled by conventional OCR.}
\label{vertVsRot}
\end{figure}

STR poses a great challenge due to the large variance of text patterns and fonts, complex foreground-background variations, imperfect imaging conditions, and highly complicated backgrounds. This makes it a much more complex task than conventional OCR. With the advent of deep learning, deep neural networks have been used to solve the task of scene text recognition. These networks are computationally expensive, aiming to achieve the best possible recognition results on popular benchmarks. Some of them also require the use of lexicon dictionaries or language-model-based corrections to improve the prediction accuracy. This increases the memory consumption of the network further. But, due to the nature of downstream applications, it becomes necessary to process images in real-time, using the limited computation power of smartphones. We aim to build an efficient and compact general-purpose text recognition system, supporting diverse symbols and characters and consequently, multiple languages.

Other than dealing with perturbations such as blurriness, contrast mismatch, altered brightness, and so on, a general-purpose OCR system must also be able to deal with variations in orientation. \cite{shi2016robust, shi2018aster} have proposed transformation modules to normalize the text image into a straight line. But vertical text, in which horizontal characters are stacked vertically, is not handled by the current OCR systems. This orientation is different from orthogonally rotated  top-to-bottom or bottom-to-top text which can be handled by clockwise or anti-clockwise rotation of the text. Here on, for the ease of understanding, we will denote the vertically stacked horizontal text as Vertical Text and the orthogonally rotated vertical text as simply Rotated Text. The distinction between these two types of images is shown in Fig. \ref{vertVsRot}. Vertical texts are predominantly found in East-Asian scripts including Chinese, Korean, Vietnamese and Japanese. Some methods recognize vertical text by detecting character boundaries and recognizing each character in the word. But, due to improper character boundaries in scene text scenarios, high latency, and less word-level context in the case of character-level recognition, this approach is not very efficient and accurate.

In this paper, we propose STRIDE: Scene Text Recognition In-Device, a light-weight CNN-LSTM based network, which performs real-time text recognition of multi-oriented, multi-scale scene text images on-device. We develop 4 models to support 4 different scripts and cumulatively support 34 languages. Each network has around 0.88M parameters, which is at least 10x smaller than existing models with comparable precision. These optimizations are covered in more detail in further sections.
The main contributions of this paper are as follows:
\begin{itemize}
\item[1] Simultaneous recognition of horizontal and vertical text
\item[2] Addition of convolution attention blocks and analyzing its impact on OCR systems
\item[3] Developing a device ready text recognition solution, with state of the art inference speed and comparable accuracy\end{itemize}

The rest of the paper is organized in the following way. Section \ref{sec:related} talks about related works. We elucidate the working of our pipeline in section \ref{sec:network}. Section \ref{sec:experiments} concentrates on the experiments we conducted and the corresponding results we achieved. The final section \ref{sec:future} takes into consideration the future improvements which can be further incorporated.

\section{Related Works}
\label{sec:related}

Though the OCR problem has received attention for many decades, the main focus was document images  \cite{nagy2000twenty}. Although OCR on document images is now well-developed, these methods fail utterly when applied to natural scene images due to a large number of variations in scene images.

In the last decade, the advent of deep learning has led to substantial advancements in STR. Earlier methods were segmentation-based methods \cite{chen2020text} that aimed to locate the characters, use character classifier to recognize the characters, and then group characters into text lines. PhotoOCR \cite{bissacco2013photoocr} was one such segmentation-based method that used a deep neural network trained on the extracted histogram of oriented gradient (HOG) features for character classification. The performance of segmentation-based methods are constrained by their dependence on accurate detection of individual characters, which is a very challenging problem. Also, most of the methods \cite{wang2012end, liu2016scene} rely on post-OCR correction, such as lexicon set or language models to capture context beyond a character.  This leads to an increase in their time and memory consumption.

Segmentation-free methods focus on mapping the entire text image to a target string. Jaderberg\cite{jaderberg2014synthetic} treated the word recognition problem as a multi-class classification problem, where each class represents a different word. But, such an approach is not scalable. 

Recent methods have construed the problem of scene text recognition as an image-based sequence recognition. CRNN \cite{shi2016end} was the first combined application of CNN and RNN for text recognition. It consisted of a fully-convolutional part of VGG \cite{simonyan2014very}, followed by 2 Bi-LSTM layers. For lexicon-free prediction, the network was trained using connectionist temporal classification (CTC) \cite{graves2006connectionist} to find the label sequence with the highest probability. 

Multiple variants have hence been proposed to improve the performance of CRNN. Image processing techniques such as background removal, super-resolution \cite{jain2020device}, and image rectification \cite{shi2016robust} have been employed to reduce the load on the downstream stages for learning complex features. Better convolutional feature extractors such as Resnet \cite{he2016deep} and RCNN \cite{lee2016recursive} have been used to better extract the text features from complex images. But these feature extractors are very deep and have high latency. 

\begin{figure*}[htbp!]
\centerline{\includegraphics[scale=0.25]{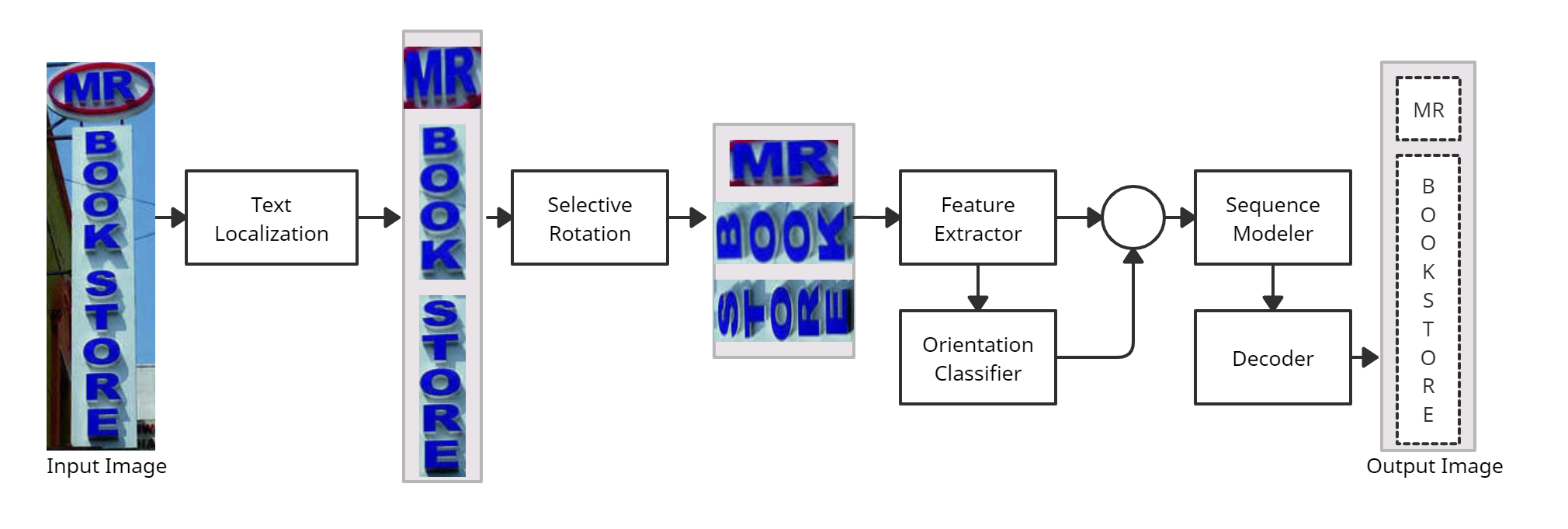}}
\caption{STRIDE Network Pipeline: The word boxes detected from the text localization network are passed to the feature extractor, after applying selective rotation. The orientation of each word is classified separately and passed to the sequence model with  the temporal word features extracted. }
\label{architecture}
\end{figure*}

Along with deep feature extractors, attention mechanism \cite{shi2016robust, cheng2017focusing, liu2016star} is often combined with RNN for character-sequence decoding.  This can help in extracting text from irregular scene text crops and achieve better performance in regular word crops too. \cite{baek2019wrong} conduct a fair comparison to identify the impact of variations in different modules. Although the newer variants have achieved better performance on various benchmarks, they have come at a cost to memory and computation. 

Convolution based Attention modules such as convolutional block attention module \cite{woo2018cbam} and global squeeze-excite blocks \cite{hu2018squeeze} have shown to increase the performance of convolutional networks on benchmarks such as Imagenet for detection and segmentation tasks. The effect of these attention modules has not been explored in the domain of text recognition till now.

The Global Squeeze-Excite blocks provide channel attention to the network and can be easily integrated with any recognition feature extraction. It recalibrates channel-wise feature responses by explicitly modeling inter-dependencies between channels. Global Squeeze-Excite blocks have been shown to increase the search space with very little effect on time and memory \cite{tan2019mnasnet}.  Alternatively, Convolutional Block Attention Module (CBAM) proposed modification to the SE block, which helps to provide spatial attention in addition to channel attention. CBAM contains two sequential sub-modules called the Channel Attention Module (CAM) and the Spatial Attention Module (SAM), which are applied in that particular order.

Most of the current STR models assume the input image as horizontal. Even the irregular scene text recognizers fail on vertical text images due to the structure of the network \cite{li2019show}. Some models use vertical information \cite{ijcai2017-458, cheng2018aon, choi2018simultaneous}. But, these models use attention mechanism, which makes them infeasible to be used on-device due to their computation requirements. 

The demand for on-device OCR has lead to prominent firms offering OCR such as Google's Mlkit\footnote{\url{developers.google.com/ml-kit/vision/text-recognition} } and Apple's Text recognition in vision framework \footnote{\url{developer.apple.com/documentation/vision/recognizing_text_in_images}}. But these products support a minimal number of languages on-device. PP-OCR \cite{du2020pp} is an open-source on-device OCR supporting multiple scripts. But, we outperform the network in terms of both accuracy and inference speed.

\section{Network Architecture}
\label{sec:network}

In this section, we describe our network, which takes an image and bounding box of the word as input and converts it into a sequence of labels. The network is based on the Convolutional Recurrent Neural Network (CRNN) architecture \cite{shi2016end}. We incorporate certain modifications to the network, to perform simultaneous recognition of horizontal and vertical text. We also focus on building a compact network, suitable for on-device inferencing without compromising on accuracy. The network architecture is shown in Fig. \ref{architecture}. The proposed model consists of four components: selective rotation, feature extraction, sequence modeling, and prediction. Each component of the network is carefully designed and optimized to minimize the overall network size.

\begin{figure*}
\centerline{\includegraphics[scale=0.22]{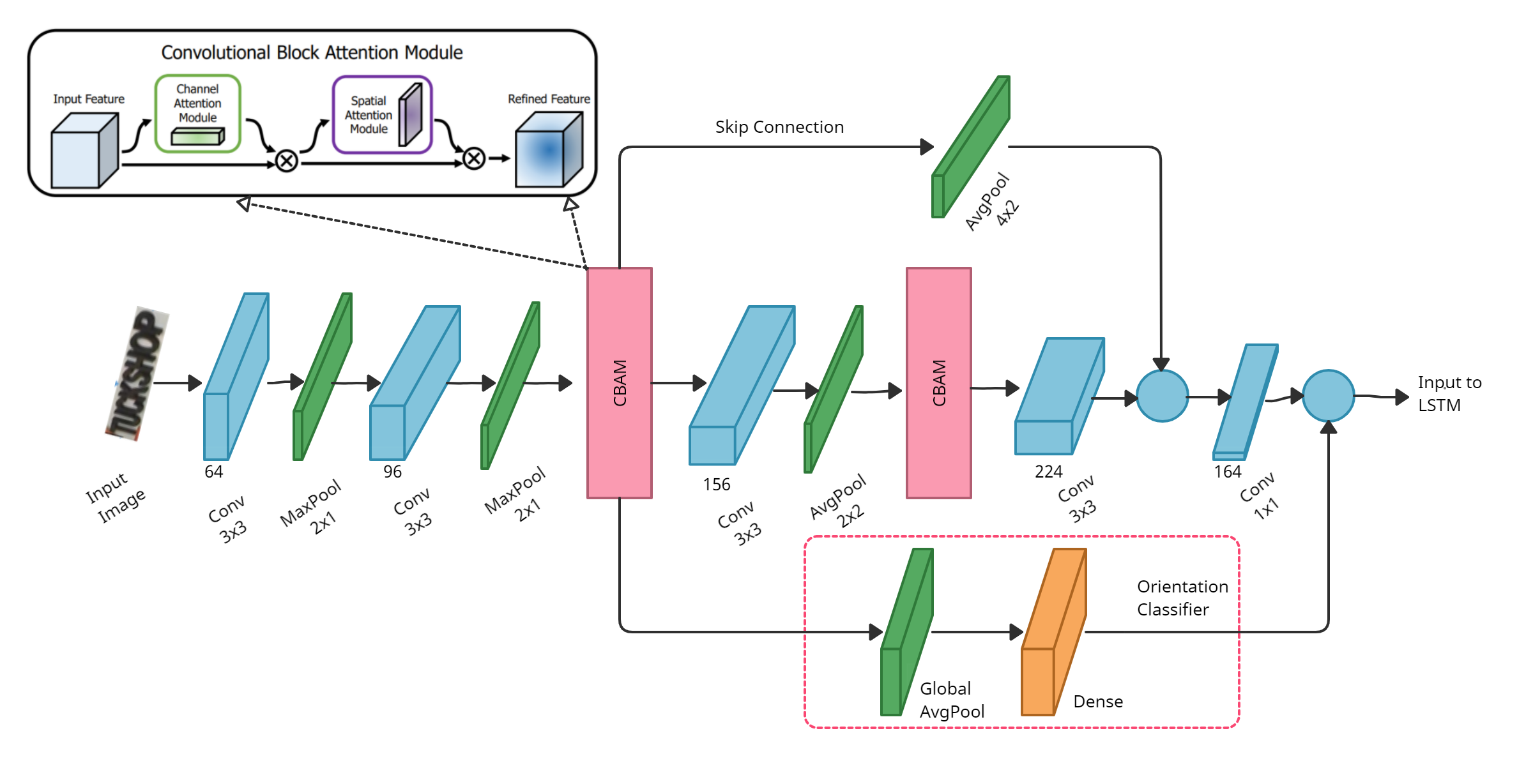}}
\caption{Feature Extractor and Orientation Classifier Module. CBAM is used to get channel and character region attention information. The detected orientation is concatenated to the extracted features and fed to the LSTM. Input Image is of size: 16*width*3, where height =16 and 3 denotes the RGB channels.  }
\label{feature_extractor}
\end{figure*}


\subsection{Selective Rotation}
Text in scene images comes in diverse shapes and varieties. The natural text present on signboards, restaurant boards, banners, etc. is generally skewed or rotated. The recognition model is made to handle perspective information, text skewness, and rotation up to a certain level. But, as the skew increases it starts impacting the recognition results. Rectifying and pre-processing each word crop adds to the computation cost of the network. To handle this trade-off between accuracy and inference time, we extract rotation angle and skewness information from the localization module \cite{telcos2021} and pass only the highly rotated words through a computer vision-based perspective transformation. The model is made invariant to small amounts of distortions in the text.

\subsection{Feature Extraction}
The feature extractor block consists of CNN and max-pooling layers which extracts attributes related to the text and learns feature representation for both horizontal and vertical text. It has 4 convolution layers with kernel size of 3x3. The first two convolutions operate on full word crop width, which is followed by an average pool layer across the width dimension. So, the features fed to the following convolution layers and the sequence model are represented by half the original word width.

Low-level features play a huge role to distinguish subtle changes in characters due to diacritics in Latin script and to correctly identify minute differences in Chinese and Korean compound characters. So, to preserve the important low-level details, we pass a residual skip-connection from the second convolution layer and concatenate it with the output of the fourth layer. All crops are scaled to a fixed height of 16 while preserving the aspect ratio. Thus, all input word-crops are of size $16*width*3$, where the $width$ of the input is determined from the aspect ratio of the original word crop.  Each convolution layer is followed by a max-pool layer across the height dimension. So, the final output from the feature extractor is of size $1*(width/2)*164$, 164 representing the number of channels fed to the sequence modeler. 

Attention mechanism provides localized information and generally boosts the recognition accuracy when used in the encoder or decoder blocks of an OCR network. But, this comes with an additional time and network complexity. \cite{baek2019wrong} shows the time gain when the attention module is used in conjunction with the base network. This additional overhead of conventional attention methods and transformer modules makes it unsuitable to be directly deployed for on-device inference. 

To overcome this drawback and take advantage of the improved feature discriminability, we propose the use of convolution attention blocks in the encoder networks for recognition. It provides attention information to the sequence modeling block with very minimal computational overhead. These blocks can be integrated with any recognition network.To the extent of our knowledge, the effect of these attention modules has not been explored in the domain of text recognition.

\begin{figure}[b!]
\centerline{\includegraphics[scale=0.65]{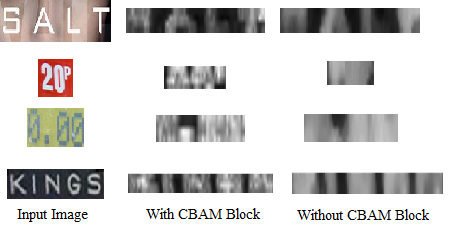}}
\caption{Feature maps extracted after the third convolution layer. CBAM blocks are able to clearly separate the characters from its background. }
\label{CBAM feature}
\end{figure}

We experiment with Global Squeeze-Excite (GSE) Blocks and Convolution Block Attention modules, both of which provide channel or region-specific localized information as described in Section 2. The GSE Blocks squeeze global spatial information into a channel descriptor and this channel attention information is mapped to the input feature. Whereas Convolution Block Attention Module (CBAM) provides both spatial and channel attention, through its two sequential sub-modules called the Channel Attention Module (CAM) and the Spatial Attention Module (SAM).

Fig. \ref{feature_extractor} shows the components of the CBAM block. In a semi-supervised way, SAM tries to learn the region of interest for each character in the word, which helps in providing a better background-foreground separation. Whereas, the CAM module focuses on learning the discriminative features between characters, that leads to excitation of desired channels.

Fig. \ref{CBAM feature} shows the feature maps from the third convolution layer of the network with and without the CBAM block. The feature maps of the network using CBAM provide better attention and character separation, which when fed to the sequence modeling block leads to better recognition results. Ablation study of SE and CBAM blocks for the text recognition model is performed in section \ref{sec:experiments}. Empirically, CBAM provides the best overall result, with minimal computational overhead.

\subsection{Orientation Classifier}
Due to the frequent use of vertical texts in East-Asian scripts including Chinese, Korean, Vietnamese, and Japanese, there is a need to recognize vertical text along with the horizontal text. 
To solve this problem, we aim to recognize horizontal and vertical text simultaneously with a single model, without any additional overhead.



 Predicting orientation at character-level is difficult due to the orthogonal nature of character pairs like Z and N or H and I. Hence, the orientation of the word requires a global word-level context, rather than a character-level context. To get this word-level information, we apply Global Average Pooling (GAP) across the width dimension. This information is then fed to a fully connected layer with sigmoid activation to predict a single orientation per word, instead of the per-pixel value of orientation across the width. Fig \ref{feature_extractor} shows the working of our orientation classifier. The prediction of the orientation classifier can be represented as:
 
 \begin{equation}
     y_{i}^{p} \ =\ \sigma \ ( \ Dense\ ( \ GAP_{axis=2}( CB_{1})))
 \end{equation}
 
 where $CB_{1}$ is the output from the first CBAM block and $y_{i}^{p}$ is the prediction of the orientation classifier. 
 
During training, we jointly optimize the recognition network and the orientation classifier. The orientation classifier is trained using binary cross-entropy loss and it is defined as:
\begin{equation}
\label{eq:orientation}
L_{o} \ =\ -\frac{1}{N} \ \sum ^{N}_{i=1} \ y_{i} \ log( y_{i}^{p}) \ +\ ( 1\ -\ y_{i}) \ log( 1\ -\ y_{i}^{p}) \ \ 
\end{equation}


where $y_{i}$ is the actual orientation, $N$ is the batch size and $L_{o}$ defines the orientation loss.

\subsection{Sequence modeling}
Sequence modeling captures contextual information within a sequence of characters, for the next stage to predict each character. Our sequence modeling stage consists of 1 Bi-LSTM layer. LSTM has shown strong capability for learning meaningful structure from an ordered sequence. Another important property of the LSTM is that the rate of changes of the internal state can be finely modulated by the recurrent weights, which contributes to its robustness against localized distortions of the input data.

The LSTM receives input from both the orientation classifier and feature extractor. The orientation information is a piece of non-temporal information that is fed to the LSTM with the temporal word features extracted from the convolution layers, the width of the image being the time dimension. There are a few ways to feed non-temporal data to the bi-directional LSTM. It can be fed as the first and last timestamp of the temporal sequence. The disadvantage of this approach is that if the input sequence is long enough, the orientation information may not be properly propagated to all the timestamps, and it might forget the conditioning data. The approach we follow is to append the orientation information to each timestamp, effectively adding the feature across the width dimension of the convolution features. We are able to retain the accuracy of horizontal text on the benchmark dataset, in addition to supporting vertical word recognition, as shown in Table \ref{tab:Vertical Text Classification}. 

For faster inference, we use LSTM with recurrent projection layer \cite{sak2014long} to decrease the computational complexity. The recurrent projection layer projects hidden states at every time step to a lower dimension. The projected hidden states in both directions are then concatenated and fed to the prediction stage.   

\subsection{Prediction}
The prediction stage outputs a sequence of characters from the identified features of the word crop. The per-frame predictions made from the LSTM are fed to a fully connected layer to obtain per-frame probability distribution over the labels. Finally, Connectionist Temporal Classification (CTC) \cite{graves2006connectionist} is used to transform frame-wise classification scores to label sequence. If $x_{t}$ is the per-frame probability distribution over the set $L'$, where $L'$ represents the set of all labels in addition to blank symbol and the ground truth label sequence is represented by $y^{*}$, then CTC defines the objective function to be minimized as follows:
\begin{equation}
    L_{c}\ =\ -\ \sum \ log\ p\left( y^{*} \ |\ x\right)
\end{equation}

Combined with orientation loss defined in Eq. \ref{eq:orientation}, the combined loss equation is:
\begin{equation}
    L\ =\ L_{c} \ + \lambda L_{o}
\end{equation}

where $\lambda$ is a hyper-parameter controlling the trade-off between the two losses.  $\lambda$ is set to 1 in our experiments. 

For fast inference, we use the greedy decoder, which assumes the most probable path to be the most probable labeling to estimate the character sequence. 

\begin{figure}[t!]
\centerline{\includegraphics[scale=1]{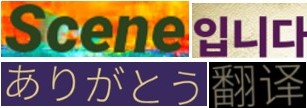}}
\caption{Synthetic dataset created by rendering font on varied backgrounds and using data augmentation techniques.}
\label{fig:synth_samples}
\end{figure}

\section{Experiments}
\label{sec:experiments}

\subsection{Datasets}
\label{subsec:datasets}

The diversity of training datasets plays an important role in creating a robust model with high performance. But, the size of the existing real-world datasets is small to  train a highly accurate scene text recognizer. The real-world vertical images are also limited in number and a few were collected from the datasets mentioned below.  Labeling scene text images is costly. Thus, we rely on synthetic datasets for training our network.\\
\textbf{HVSynth}: We create our own synthetic dataset for horizontal and vertical text, by generating text images with basic data augmentation techniques such as rotation, perspective distortion, blurring, etc. We also apply text blending, to add background noise to the text. We create a total of 5L samples for horizontal text and 1L for vertical text. Fig. \ref{fig:synth_samples} and Fig. \ref{vertical_Examples} shows some samples of synthetic horizontal and vertical text.

\begin{figure}[t!]
\centerline{\includegraphics[scale=0.65]{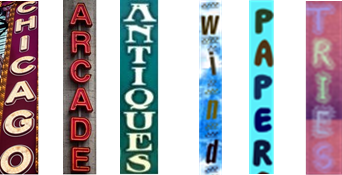}}
\caption{Vertical Natural and Synthetic Text Samples }
\label{vertical_Examples}
\end{figure}


Along with our synthetic datasets, we use standard synthetic datasets such as: \\
\textbf{MJSynth} (MJ) \cite{jaderberg2014synthetic} is a synthetic dataset designed for Scene Text Recognition. It contains 8.9M word box images generated using 90k alpha-numeric words. We use this dataset to train our Latin model.   \\
\textbf{SynthText} (ST) \cite{gupta2016synthetic} is another synthetically generated dataset. Even though SynthText was an arrangement for detection, it has been also used for Scene Text Recognition by cropping off word boxes. SynthText has around 6M training data once we render the text samples as word boxes. We also use the same method to generate word crops for all the other scripts.

We use the following real-world datasets for training and evaluation:\\
\textbf{ICDAR 2013 (IC13)}  \cite{karatzas2013icdar}  was created for the ICDAR 2013 Robust Reading competition for reading camera captured scene texts. The dataset has 848 images for training and 1095 images for testing. \\
\textbf{ICDAR 2015 Incidental Text (IC15)} \cite{karatzas2015icdar} consists of a lot of irregular text and highly blurred images. It has 4468 training images and 2077 testing images. \\
\textbf{IIIT5k-Words (IIIT5k)} \cite{mishra2012scene} consists of images crawled from google search result. The dataset consists of 5000 images that are split into a training set of 2000 images and an evaluation set of 3000 images. \\
\textbf{Street View Text (SVT)} \cite{wang2011end} consists of 247 images for training and 647 images for evaluation. The images are outdoor street images collected using Google Street view. Some images are severely corrupted by noise, blur, and low resolution.\\
\textbf{ICDAR 2019 multi-lingual scene text (MLT) } \cite{nayef2017icdar2017} consists of around 90k word crops belonging to 7 scripts. Though the dataset was created for script identification, the word crops have been annotated and can be used for training STR. 


\subsection{Training details}
 We implement the network using Tensorflow 2.3 \cite{abadi2016tensorflow} framework. All experiments are carried out on  GeForce GTX 1080ti GPU, and 16GB RAM. The training batch size is 64. We use Adam optimizer \cite{kingma2014adam} to train the models with initial learning rate $1e-3$. The learning rate was halved when the validation loss did not decrease for more than 2 epochs, and the training was stopped once the learning rate reached $1e-5$.  
 
 Training CTC is hard and the models take a lot of time to converge \cite{borisyuk2018rosetta}. For improving stability during training, we explore curriculum learning strategies. Curriculum learning was proposed to address the challenges of training deep neural networks under non-convex training criteria. For a few epochs, we train our network on a subset of training images consisting of shorter words and clear text. After every epoch, we increased the maximum length of the words along with the degree of perturbations. Adopting curriculum training strategies significantly fastened the training of the network.

Both synthetic and natural scene text contains few wrongly annotated images and highly complex background images. Also, many images of the MjSynth, SynthText, and IIIT5k datasets have case-insensitive labeling, which makes it difficult for the model to learn case information. To tackle these data-related issues, we weakly supervise our model, by monitoring per image loss. We reduce the training data instances that have a very high loss on a trained model, which could possibly be due to wrong annotations or hard examples. We also correct the wrong case annotations of the dataset, by taking input from the model, in case of incorrect case predictions. The dropped images are fed back to the data pipeline after a few epochs, to make up for the dropped hard examples.

\subsection{Results on benchmark datasets}

Though there are a few benchmark recognition datasets for Latin, there are no such datasets for other scripts. There are end-to-end benchmarks for Chinese, but none for recognition.
The recognition word accuracies on the public real-word datasets mentioned in section \ref{subsec:datasets}, obtained by some state-of-the-art techniques and our proposed model STRIDE are given in Table \ref{tab:Experimental Results}. The results of the other models are taken from \cite{baek2019wrong} which compares different STR models.

\begin{table}[t!]
\label{tab:Experimental Results}
\centering
\caption{Experimental Results}\label{tab1}
\begin{tabular}{| M{1.35cm} |  M{1cm} | M{0.9cm} | M{0.9cm} | M{0.9cm} | M{0.9cm} | M{0.9cm} | }
\hline
\bfseries {Model} & 
\bfseries{Params} &
\bfseries{IIIT} & 
\bfseries{SVT} & 
\bfseries{IC13} & 
\bfseries{IC15} \\
\hline
CRNN & 8.3M &  82.9 & 81.6 & 89.2 & 64.2   \\
RARE &  10.8M & 86.2 & 85.8 & 91.1 & 68.9 \\
STAR-Net & 48.7M & 87.0 & 86.9 & 91.5 & 70.3  \\
Rosetta & 44.3M & 84.3 & 84.7 & 89.0 & 66.0 \\
Our Model & 0.88M  &  82.5 & 81.3 & 88.4 & 68.7 \\
\hline
\end{tabular}
\label{tab:Experimental Results}
\end{table}

Though our network is a minuscule version of CRNN having around a tenth of its parameters, the recognition accuracies of our model is on-par with CRNN. We perform better in IC15, since it consists of a few vertical and orthogonally rotated images which our model can handle. By using computationally intensive modules such as attention for decoding (in RARE) and Resnet for feature extraction (in Star-Net), the performance gap widens. But due to the on-device constraints, we cannot integrate these modules in our model.

\begin{table}[b!]
\centering
\caption{Vertical Text Classification and Recognition Result on Latin synthetic dataset}\label{tab1}
\begin{tabular}{| M{3.0cm} | M{1.25cm} | M{1.25cm} | M{1.25cm} |}
\hline
\bfseries {Network} & 
\bfseries{Horizontal Word Accuracy} & 
\bfseries{Vertical Word Accuracy} & 
\bfseries{Orientation Classifier Accuracy}\\
\hline
Base & 94.77 & - & - \\
Base + Orientation Classifier &94.25 & 96.12 & 97.34 \\

\hline
\end{tabular}
\label{tab:Vertical Text Classification}
\end{table}

\subsection{Result on HVSynth dataset}
The performance of our model on the HVSynth dataset is shown in Table \ref{tab:Vertical Text Classification}. Our model is able to predict well on both horizontal and vertical word crops with a negligible loss on horizontal word crops performance, compared to our base model. The orientation classifier is able to learn the distinction between horizontal and vertical text and predicts the orientation with high accuracy. 

\begin{table}[b!]
\label{tab:Ablation Study of GSE and CBAM modules on IC13 data}
\centering
\caption{Ablation Study of GSE and CBAM modules on IC13 data}\label{tab:attention}
\begin{tabular}{| M{1.85cm} | M{1.2cm} | M{1.2cm} | M{1.25cm} | M{1.08cm} |}
\hline
\bfseries {Module} & 
\bfseries{Word Accuracy} & 
\bfseries{Char Accuracy} & 
\bfseries{Parameters} &
\bfseries{Time}\\
\hline
Base Network & 86.5 & 92.2 & 850k & 2.2 ms \\
1 GSE Block & 87.4 & 92.8 & 859k & 2.28 ms\\
2 GSE Block & 87.7 & 93.1 & 868K & 2.35 ms\\
2 CBAM Block & 88.4 & 93.4 & 886K & 2.44 ms \\
\hline
\end{tabular}
\label{tab:Ablation Study of GSE and CBAM modules on IC13 data}
\end{table}

\subsection{Ablation study on attention modules}

Table \ref{tab:attention} shows the impact of attention modules on the recognition accuracies of our Latin model on IC13. As we can see, convolutional attention blocks can boost the performance of the network with little effect on latency and size. Thus, using CBAM, which provides both spatial and channel attention, helps in better extraction of the features of the characters.

\begin{figure}[t!]
\centerline{\includegraphics[scale=0.65]{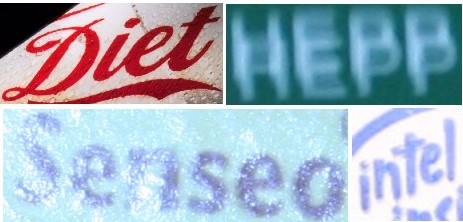}}
\caption{Complex Pass Cases}
\label{Pass Cases}
\end{figure}

\subsection{Results Analysis}
Our model performs substantially well while handling complex images like those shown in Fig. \ref{Pass Cases}. We also  investigate the failure cases of our model and some of them are shown in Fig. \ref{Fail_Cases}. A few common reasons for failure are elucidated below:

\begin{figure}[b!]
\centerline{\includegraphics[scale=0.45]{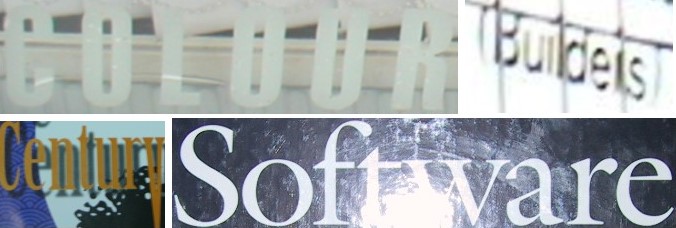}}
\caption{Fail Cases}
\label{Fail_Cases}
\end{figure}

\begin{itemize}
\item \textbf{Text Merging with Background:} Model fails to handle cases where the text merges with the background. Better feature extractors may improve performance in this domain.

\item \textbf{Shadow and Blur:} Selective failure cases are observed due to reflections in low resolution textured images.


\item \textbf{Smudged Text:} Existing models do not explicitly handle smudged text in low resolution cases; super-resolution modules or image pyramids might improve performance.

\item \textbf{Occluded Text:} Current research methods do not substantially exploit contextual information to overcome occlusion. Future researches may utilize superior language models to maximally make use of context.

\end{itemize}

\subsection{On-Device Statistics}
The model details for various scripts are shown in Table \ref{tab:ondevice}. The number of parameters of the model varies across scripts due to the difference in the number of characters supported, which affects the number of parameters in the last fully connected layer. For speed comparisons, we find out the time taken to process a word crop of size $16*64$ on S20 which has Exynos 990 chipset and 8 GB RAM. 

\begin{table}
\centering
\caption{On-Device Statistics}
\label{tab:ondevice}
\begin{tabular}{| M{1.0cm} | M{2.0cm} | M{2.0cm} | M{2.0cm} |}
\hline
\bfseries {Script} & 
\bfseries{No. of characters supported } & 
\bfseries{No. of parameters (in mil)} & 
\bfseries{Time taken (in ms)}\\
\hline
Latin & 236 & 0.88 & 2.44   \\
Korean & 1330 & 0.99 & 2.81 \\
Japanese & 2647 & 1.11  & 3.21 \\
Chinese & 5949 & 1.43 & 3.93\\
\hline
\end{tabular}
\end{table}

\section{Future Work}
\label{sec:future}
Though the model performs well on regular datasets, the model does not perform well on irregular datasets containing curved text and distorted text due to model limitations. \cite{baek2019wrong} has clearly shown that such images can be deciphered by using computationally expensive modules such as 2-D attention \cite{li2019show}. Other solutions to read irregular word crops involve developing efficient pre-processing modules that can transform and normalize the image \cite{shi2016robust}. We plan to explore both avenues to develop a robust on-device OCR that can handle both types of word crops.

Another interesting scope of future work will be to target challenging scripts like Arabic which are written right to left. The model also faces difficulty in extracting text written in calligraphic fonts and handwritten text. The model can be fine-tuned with such datasets and it remains to be seen how this similar architecture will perform in such cases.

\section{Conclusion}
In this paper, we have presented STRIDE, a lightweight OCR solution. 
OCR is one of the most important and popular areas of research especially for scene text given the popularity of high-resolution cameras in recent times, and the billions of images captured daily. In this paper, we presented STRIDE, our novel, lightweight, lexicon-free, and on-device solution with real-time results. We have demonstrated that the architecture of our proposed system is universal, and scales to document and scene text, and to other scripts and languages as well. The system can be used to recognize both horizontally and vertically aligned text. We also introduce the use of convolution attention blocks in STR networks and show its impact on accuracy through ablation studies. Additionally, we have compared our methods to previous works. We have shown how we improve upon previous approaches and handle the complex scenarios and challenges of scene text, at the same time optimizing the network for real-time performance on-device. The number of LSTM layers is decided to achieve a trade-off between model performance and model size. Stacking more LSTM layers didn't give any significant boost in model accuracy. We are able to achieve comparable results with the current SOTA recognition models with only 0.88M parameters and an On-Device inference time of 2.44ms. We hope this will enable the use of OCR on camera images and scene text as a starting point for various downstream Vision and NLP tasks, entirely on-device giving real-time results and protecting the user's privacy at the same time.

\bibliographystyle{IEEEtran}
\bibliography{IEEEabrv,references}

\end{document}